\documentclass[runningheads]{llncs}
\usepackage[T1]{fontenc}
\usepackage[T1]{fontenc}
\def\doi#1{\href{https://doi.org/\detokenize{#1}}{\url{https://doi.org/\detokenize{#1}}}}

\usepackage{graphicx}
\usepackage{listings}
\usepackage{hyperref}
\usepackage{subfigure}
\usepackage{textcomp}
\usepackage{verbatim}
\usepackage{multirow}
\usepackage{multicol}
\usepackage{color}
\usepackage{xcolor}
\usepackage{bbm}
\usepackage{amssymb}
\usepackage{marvosym}

\usepackage[linesnumbered, lined, boxed, commentsnumbered, ruled, vlined]{algorithm2e}
\usepackage{tabularx,booktabs}

\usepackage{pifont}
\newcommand{\smalltitle}[1]{{\noindent\textbf{#1.}\hspace{1mm}}}

\begin{document}

\title{Self-accumulative Vision Transformer for Bone Age Assessment Using the Sauvegrain Method}
\author{Hong-Jun Choi\textsuperscript{(\Letter)}\inst{1}, Dongbin Na\inst{1}, Kyungjin Cho\inst{1,2}, Byunguk Bae\inst{1}, Seo Taek Kong\inst{1,3}, Hyunjoon An\textsuperscript{(\Letter)}\inst{1}}

\institute{VUNO Inc., Seoul, South Korea\\ 
\email{\{hongjun.choi, hyunjoon.an\}@vuno.co}\\
\and
University of Ulsan, College of Medicine
Asan Medical Center, Seoul, South Korea\\\and
University of Illinois at Urbana-Champaign, Urbana, IL, USA
}

\authorrunning{Choi et al.}
\titlerunning{Self-Accumulative Vision Transformer}
\maketitle 

\begin{abstract}
This study presents a novel approach to bone age assessment (BAA) using a multi-view, multi-task classification model based on the Sauvegrain method.
A straightforward solution to automating the Sauvegrain method, which assesses a maturity score for each landmark in the elbow and predicts the bone age, is to train classifiers independently to score each region of interest (RoI), but this approach limits the accessible information to local morphologies and increases computational costs.
As a result, this work proposes a self-accumulative vision transformer (SAT) that mitigates anisotropic behavior, which usually occurs in multi-view, multi-task problems and limits the effectiveness of a vision transformer, by applying token replay and regional attention bias.
A number of experiments show that SAT successfully exploits the relationships between landmarks and learns global morphological features, resulting in a mean absolute error of BAA that is 0.11 lower than that of the previous work. 
Additionally, the proposed SAT has four times reduced parameters than an ensemble of individual classifiers of the previous work.
Lastly, this work also provides informative implications for clinical practice, improving the accuracy and efficiency of BAA in diagnosing abnormal growth in adolescents.
\end{abstract}

\section{Introduction}\label{sec:introduction}
 \begin{figure*}[t]
	\centering 
 \subfigure[\label{fig:landmark} ]
    {\hspace{1mm}\includegraphics[width=0.58\columnwidth]{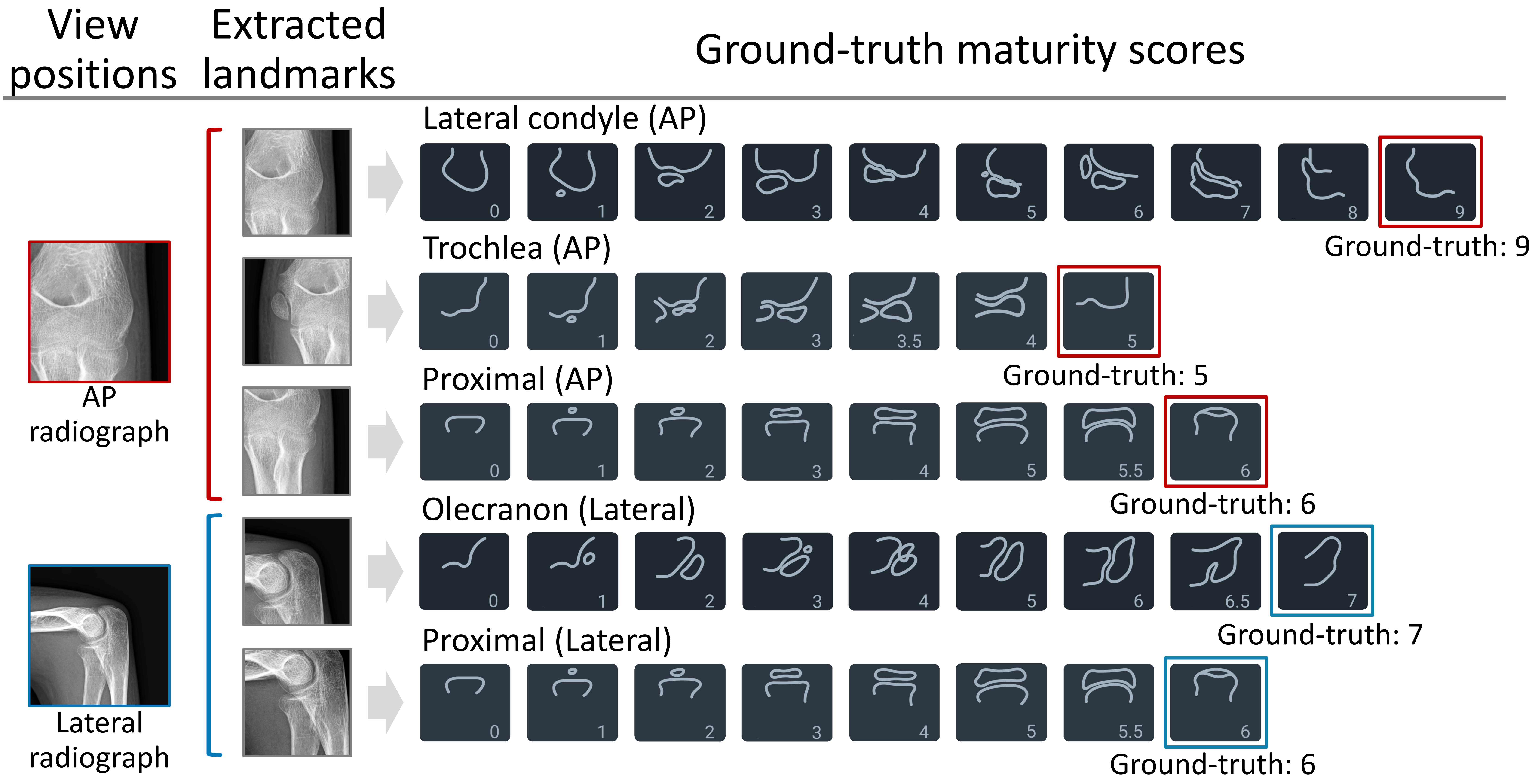}\hspace{1mm}}
     \subfigure[\label{fig:corr} ]{\hspace{1mm}\includegraphics[width=0.37\columnwidth]{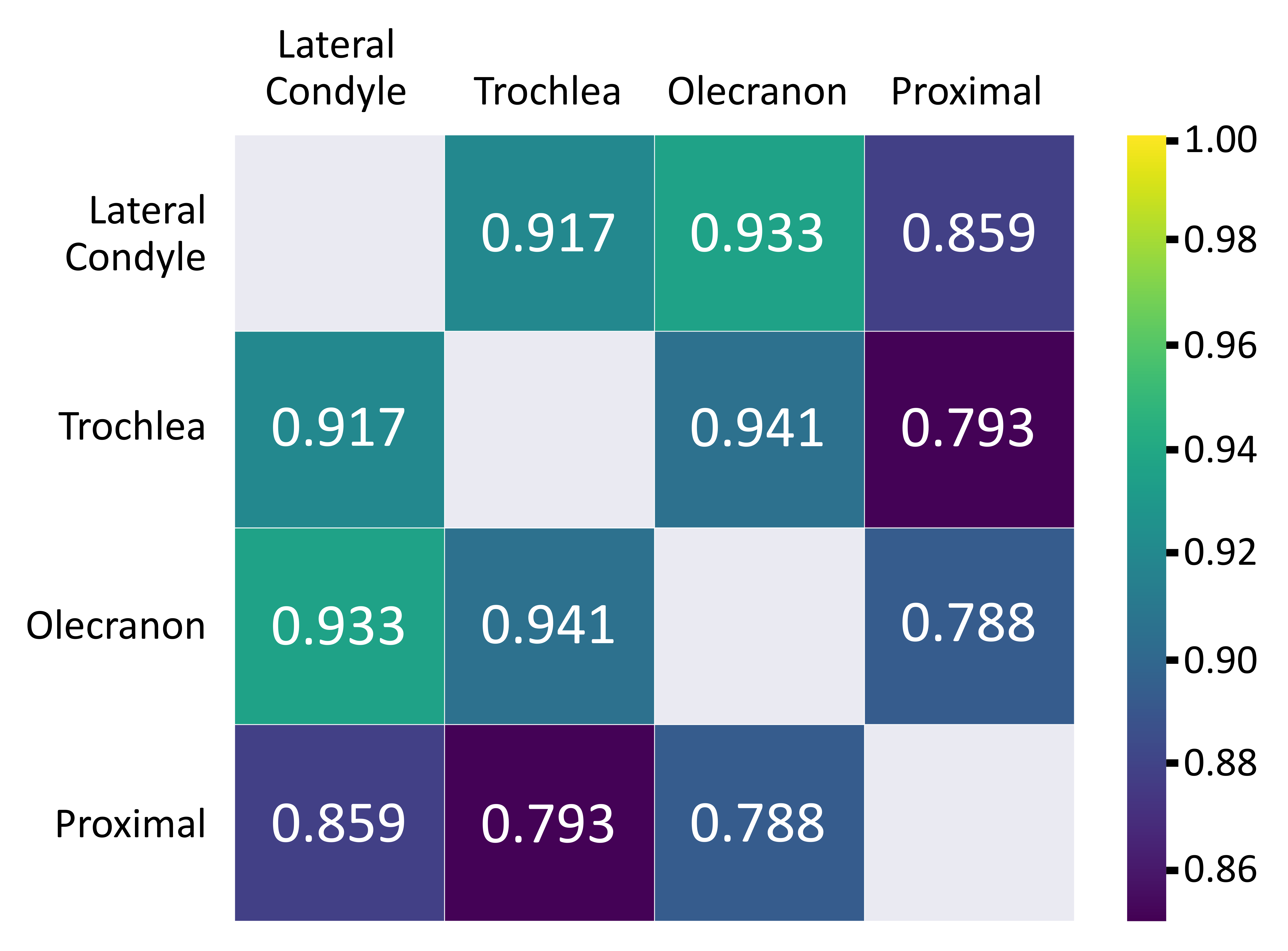}\hspace{1mm}}
     \caption{(a) An example of labeling procedure for elbow radiographs in the anterior posterior (AP) and lateral view points. The total score is calculated by summing each maturity score, with the expectation of two proximal landmarks. In this case, the total score is calculated as 9 + 5 + (6 + 6) / 2 + 7 = 27. (b) Pearson correlation coefficient of inter-landmark maturity scores.
    }
\end{figure*}

Bone age assessment (BAA) is widely used to diagnose precocious or delayed puberty.
Deep neural network (DNN) has demonstrated remarkable success in a wide range of safety-critical applications, and it also has been actively adopted in BAA with successful applications to computer-aided diagnosis systems~\cite{lee2017fully,spampinato2017deep}.
To develop an application that predicts bone age, the DNN can be trained with several clinical criteria, including Greulich-Pyle \cite{greulich1959radiographic}, Tanner-Whitehouse \cite{tanner2001assessment}, and Sauvegrain method \cite{sauvegrain1962study}.
Among these, the Sauvegrain method, which evaluates the skeletal age based on four elbow landmarks (i.e., lateral condyle, trochlea, proximal, olecranon) from two different views (see Fig \ref{fig:landmark}), is more adequate for the age of puberty \cite{ahn2021assessment,dimeglio2005accuracy}.
Specifically, the Sauvegrain method has an important trait in that the relationship among landmarks has a correlation on the label space (see Fig. \ref{fig:corr}).

A recent study has used deep learning algorithms to apply the Sauvegrain method by training an individual convolutional neural network (CNN) to assess the maturity point for each landmark \cite{ahn2021assessment}.
This approach involves each classification model inferring a score for a region of interest (RoI), and the aggregate of scores from multiple models is used to determine skeletal age.

Although the previous study shows that the classification performance of the CNNs is comparable to experts, their approach has two limitations;
First, while ground-truth labels between inter-landmarks have a strong correlation, the incorporated model that each classifier trained with single landmark images can produce misclassified predictions with high variance. These inaccurate predictions may confuse radiologists when interpreting the model’s decisions.
Second, the previous method requires excessive computational costs both training and inference because multiple landmark networks should be trained independently.

To address the above issues, this study poses a novel approach to solving the multi-view and multi-task problem for Sauvegrain-based BAA using a vision transformer (ViT) \cite{vit} instead of an ensemble of single-view CNNs.
By leveraging an attention mechanism in ViT, the model learns effective relations within input sequences which consist of multi-view inputs.
Moreover, we can reduce the number of parameters and computational costs by adopting shallow RoI-specific classifiers at the top of the shared encoder.
Although ViT has been applied to multi-view \cite{mvt,neimark2021video,sun2022transformer} and multi-task \cite{multitask_vit} problems, but not when they coexist.

However, we find that the vanilla ViT trained with a multi-view and multi-task (MV-MT) manner suffers from poor optimization and generalization. 
One of the reasons is that inter-landmarks are often excessively accentuated.
As a result, anisotropic behavior in the attention layer leads to sub-par classification performance.
To overcome the above challenge, we propose the \textit{self-accumulative vision transformer} (SAT) that accumulates their intra-region information by two components: (1) \textit{token replay} that prevents semantic representations of tokens with the same landmark from being overwhelmed by other regional tokens by using residual connections between class tokens and their corresponding regional tokens, and (2) \textit{regional attention bias (RAB)}, modified self-attention mechanism, to impose an intra-region attention.
Despite having significantly fewer parameters, the proposed SAT predicts maturity scores across landmarks much more accurately and outperforms other state-of-the-art models on most landmarks.
\section{Method}\label{sec:method}

\begin{figure*}[t]
	\centering 
     \subfigure[\label{fig:overview} ]
    {\includegraphics[width=0.35\columnwidth]{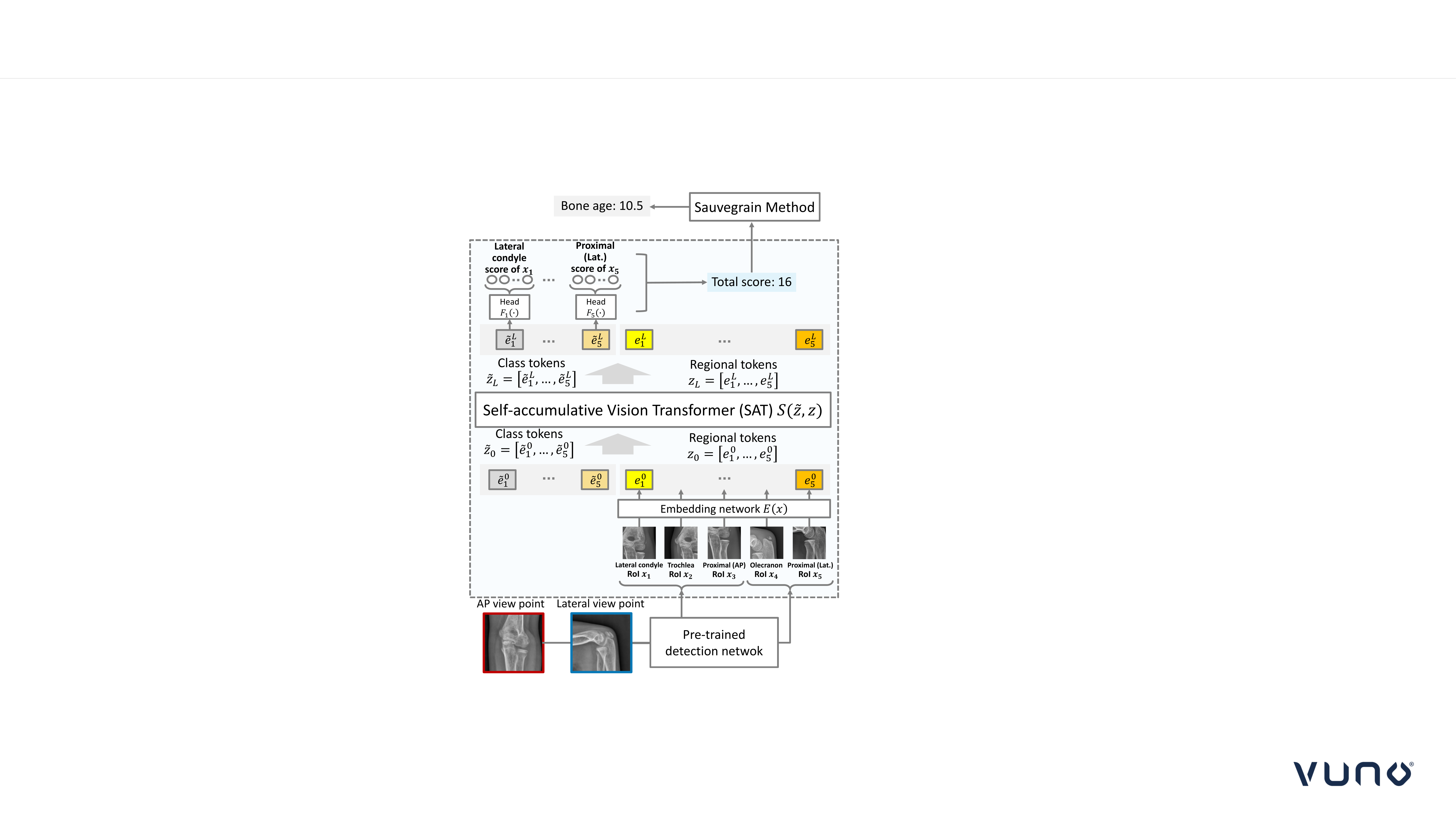}\hspace{1.5mm}}
    \subfigure[\label{fig:dongbin_figure_5} ]
    {\includegraphics[width=0.24\columnwidth]{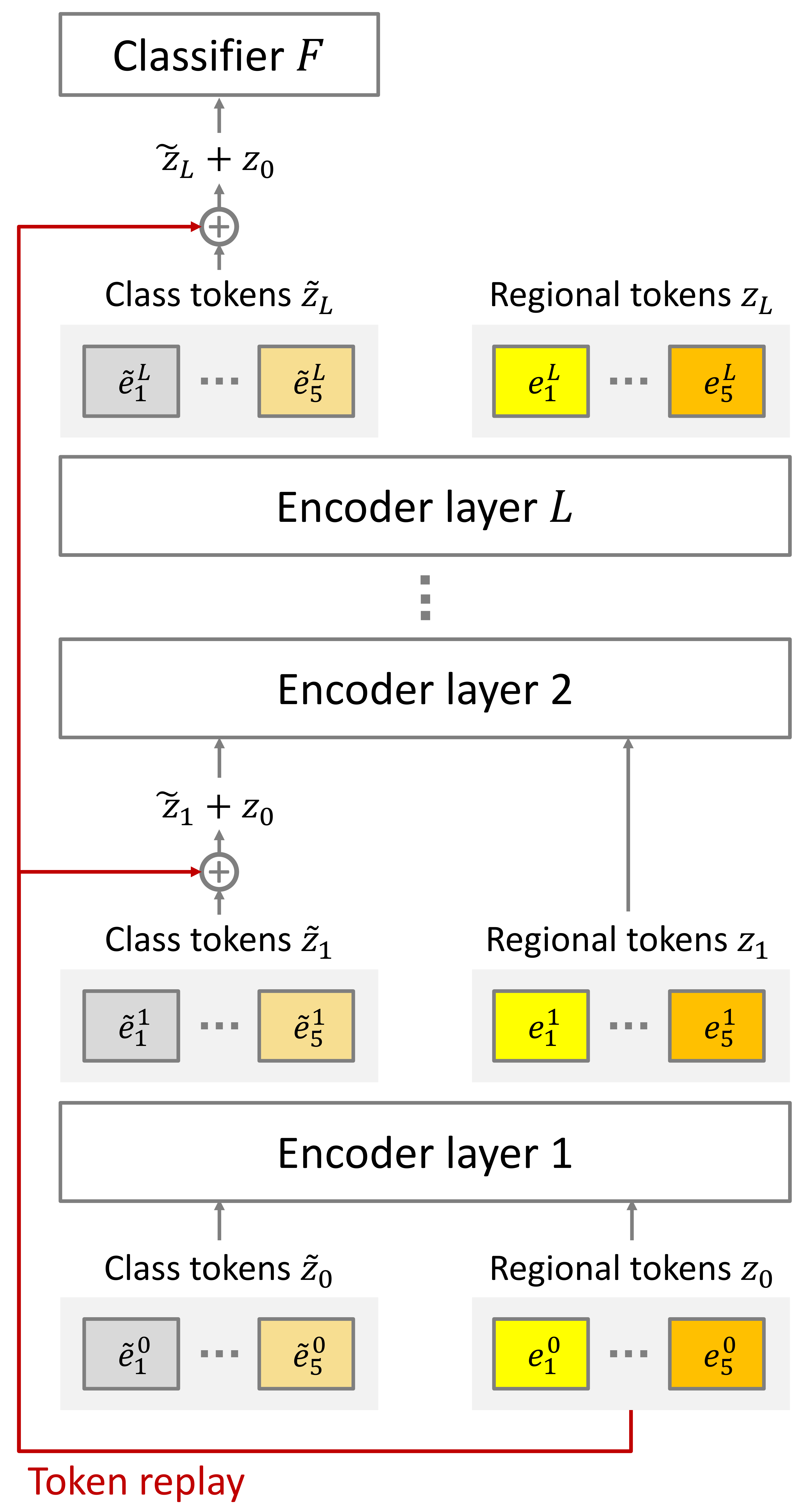}}
    \subfigure[\label{fig:dongbin_figure_6} ]
    {\includegraphics[width=0.3\columnwidth]
    {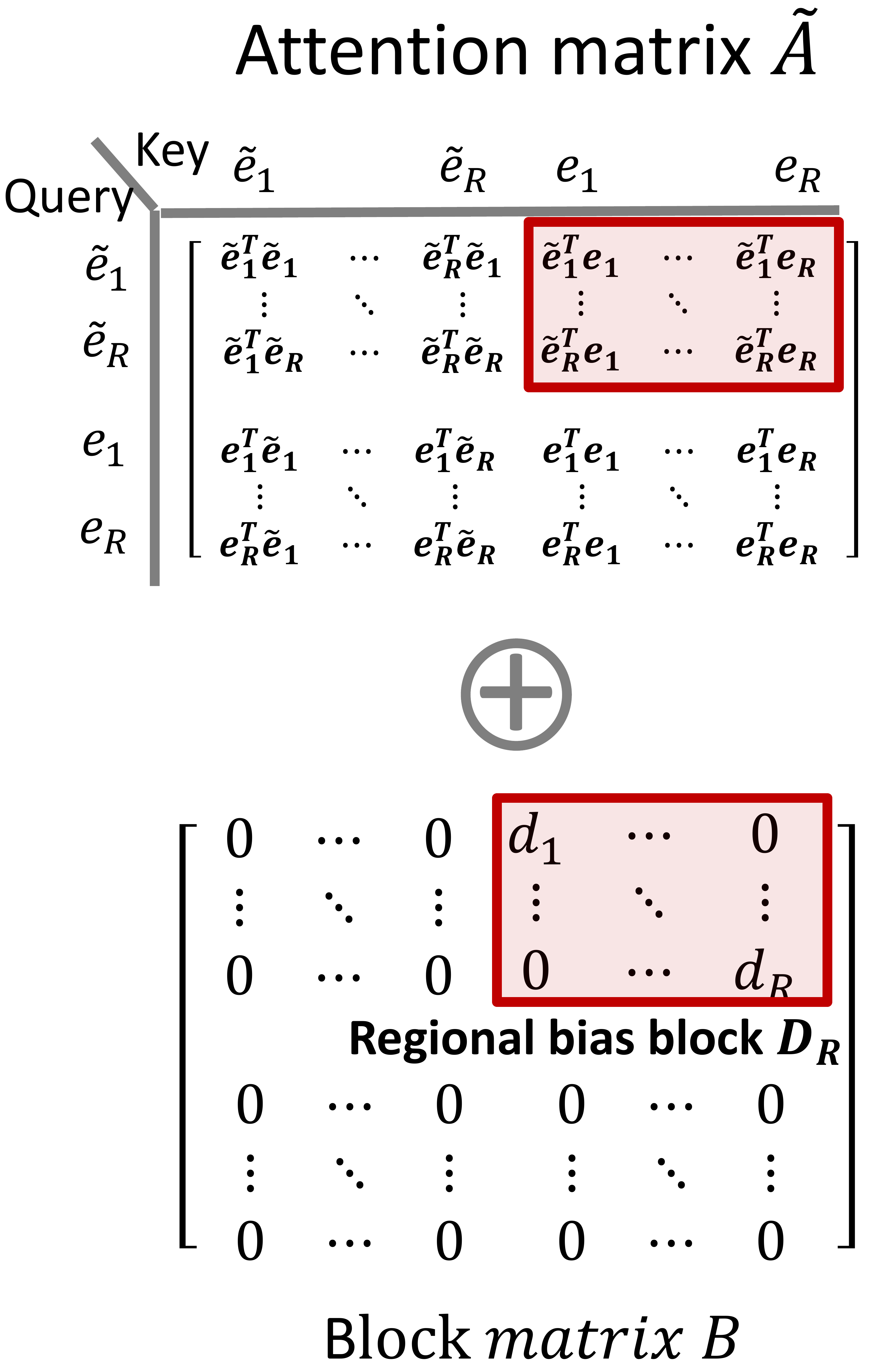}}
    \caption{(a) An overview of the Sauvegrain method with SAT. (b) Token replay prevents semantic representation from vanishing by adding the regional tokens $z_0$ to class tokens from $\tilde{z}_1$ to $\tilde{z}_L$. 
    (c) Block matrix $B$ is added on attention matrix $\tilde{A}$. Thus regional bias block $D_R$, where diagonal values are RAB of each region $r$, is added to the top-right side of the attention matrix $\tilde{A}$. 
    }
    \label{fig:dongbin_figure_temp}
\end{figure*}

\subsection{Preliminary}
\subsubsection{MV-MT Vision Transformer.}
Here we describe how we train the vanilla MV-MT ViT as its variant for ordinal classification.
We adopt a hybrid vision transformer where a patch-wise projection module is replaced by a CNN encoder with an average pooling $E: \mathcal{X} \rightarrow \mathbb{R}^d$ (e.g., ResNet~\cite{resnet})
Therefore, a RoI image $x \in \mathbb{R}^{H \times W \times C}$ is processed with the embedding network: $e_r^0 = E(x_r)$, where $r \in \{1,...,R\}$ indicates the landmark index and $R=5$.

Before feeding them to the ViT, learnable \verb|[CLS]| tokens $\tilde{z}_0 := \left[\tilde{e}_1^0, \dots, \tilde{e}_R^0\right]$ are prepended to the embedded sequence of regional tokens $z_0 := \left[e_1^0, \dots, e_R^0\right]$.
A sequence of tokens $\left[\tilde{z}_0, z_{0}\right] \in \mathbb{R}^{2R \times d}$ is then fed to the ViT with $L$ encoder layers $\{h_l\}_{l=1}^L$.
For obtaining $R$ RoI predictions, we use the final \verb|[CLS]| tokens $\tilde{z}_L$ to classify the maturity scores with each RoI-specific classifier which consists single dense layer.
The detailed comparison illustrations for architectures between SV-ST, MV-ST, and MV-MT are shown in the supplementary section.

\noindent \textbf{Ordinal Classification}. To estimate the bone age of an individual, where their classes have an ordered relation, our method handles the ordinal classification.
Therefore, we adopt the mean-variance loss~\cite{mvloss} as our loss function.
In our framework, which addresses multi-view and multi-task, each region $r$ is associated with different numbers of classes (scores) $K_r$.
The maturity score probability distribution $p_r$ of the region $r$ is calculated by forwarding an embedding vector $\tilde{e}_r^L$ introduced by the last $L$-th encoder layer into a classification head $F_r(\tilde{e}_r^L)$.
Consequently, we can get the probability value for $k$-th label of the region $r$ as $p_{r,k}$ ($k\in\{1,2,...,K_r\}$).
Given a predicted probability $p_{r,k}$ value over $K_r$ possible scores and $y_r$ its ground-truth label, the mean loss $\mathcal{L}_{\mu}(x)$ for an region image $x$ is defined as:
\begin{equation}
    \label{eq:mean}
    \mathcal{L}_{\mu}(x) = \frac{1}{R} \sum_{r=1}^{R}(\mu_r - y_r)^2 = \frac{1}{R}\sum_{r=1}^{R}(\sum_{k=1}^{K_r}k*p_{r,k}-y_r)^2.
\end{equation}
We utilize the mean squared error (MSE) loss for reducing the difference between predicted mean $\mu_r=\mathbb{E}_{\hat{y} \sim p_r}\left[\hat{y}\right]$ and the underlying ground-truth score $y_r$.
Similarly, the variance loss $\mathcal{L}_{\sigma^2}(x)$ for an region image $x$ is defined as:
\begin{equation}\label{eq:var}
    \mathcal{L}_{\sigma^2}(x) = \frac{1}{R}\sum_{r=1}^{R}\sum_{k=1}^{K_r}p_{r,k}*(k-\mu_r)^2.
\end{equation}
Thus, considering the dataset size of $N$, mean loss and variance loss is calculated as $\frac{1}{N}\sum_{i=1}^{N}\mathcal{L}_{\mu}(x_i)$ and $\frac{1}{N}\sum_{i=1}^{N}\mathcal{L}_{\sigma^2}(x_i)$ respectively.
Finally our model is optimized by following total loss:
\begin{equation}\label{eq:total}
    \mathcal{L}_{total}(x) = \mathcal{L}_{ce}(x) + \lambda_{\mu} \cdot \mathcal{L}_{\mu}(x) +\lambda_{\sigma^2} \cdot \mathcal{L}_{\sigma^2}(x),
\end{equation}
where $\mathcal{L}_{ce}$ is the cross-entropy loss, coefficient $\lambda_{\mu}$ and $\lambda_{\sigma^2}$ is a hyperparameter to adjust the weight of each loss function. In our work, we have found that $\lambda_{\mu}$ and $\lambda_{\sigma^2}$ works best at 0.2 and 0.05 respectively.

\subsection{Analysis on Anisotropic Relations between Landmarks}
\begin{figure*}[h]
    \centering 
    \subfigure[\label{fig:relevance}]{\hspace{0mm}\includegraphics[width=0.54\columnwidth]{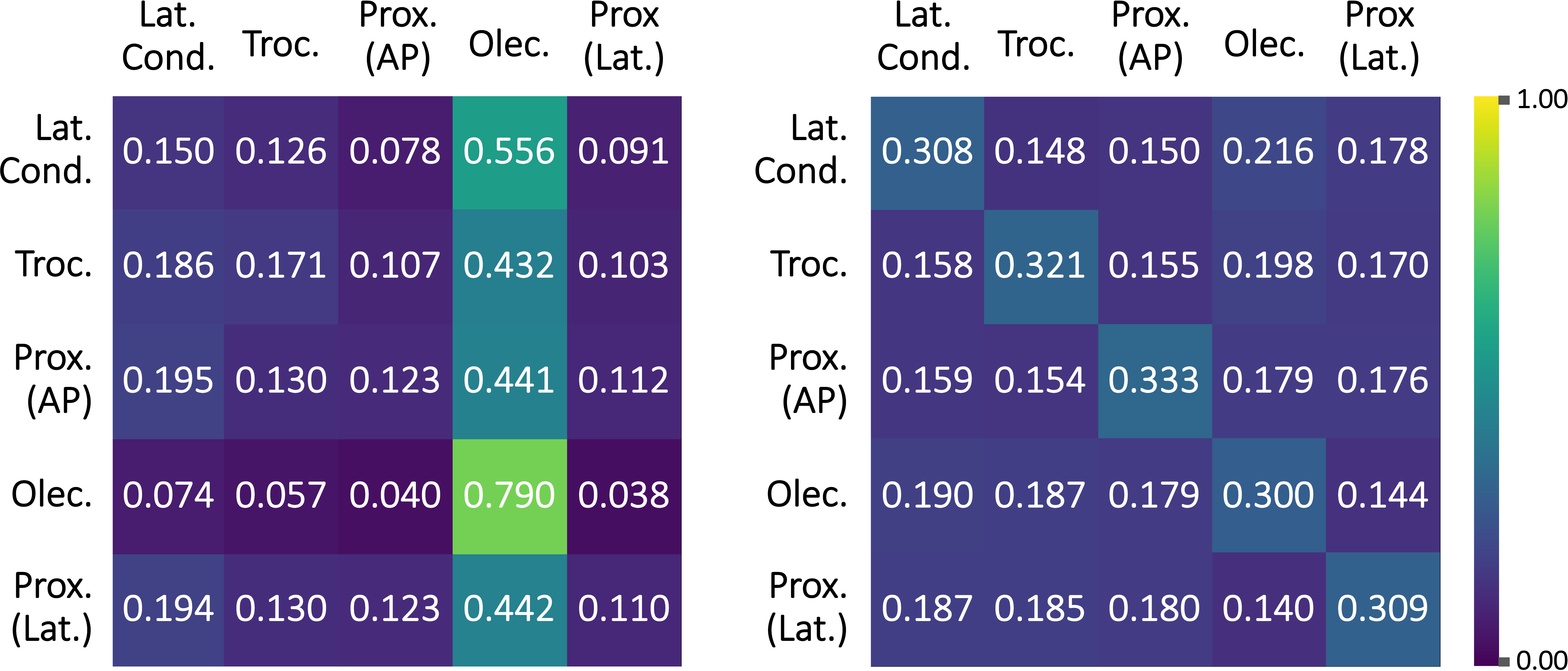}\hspace{0mm}}
    \subfigure[\label{fig:learning_curve}]{\hspace{0mm}\includegraphics[width=0.44\columnwidth]{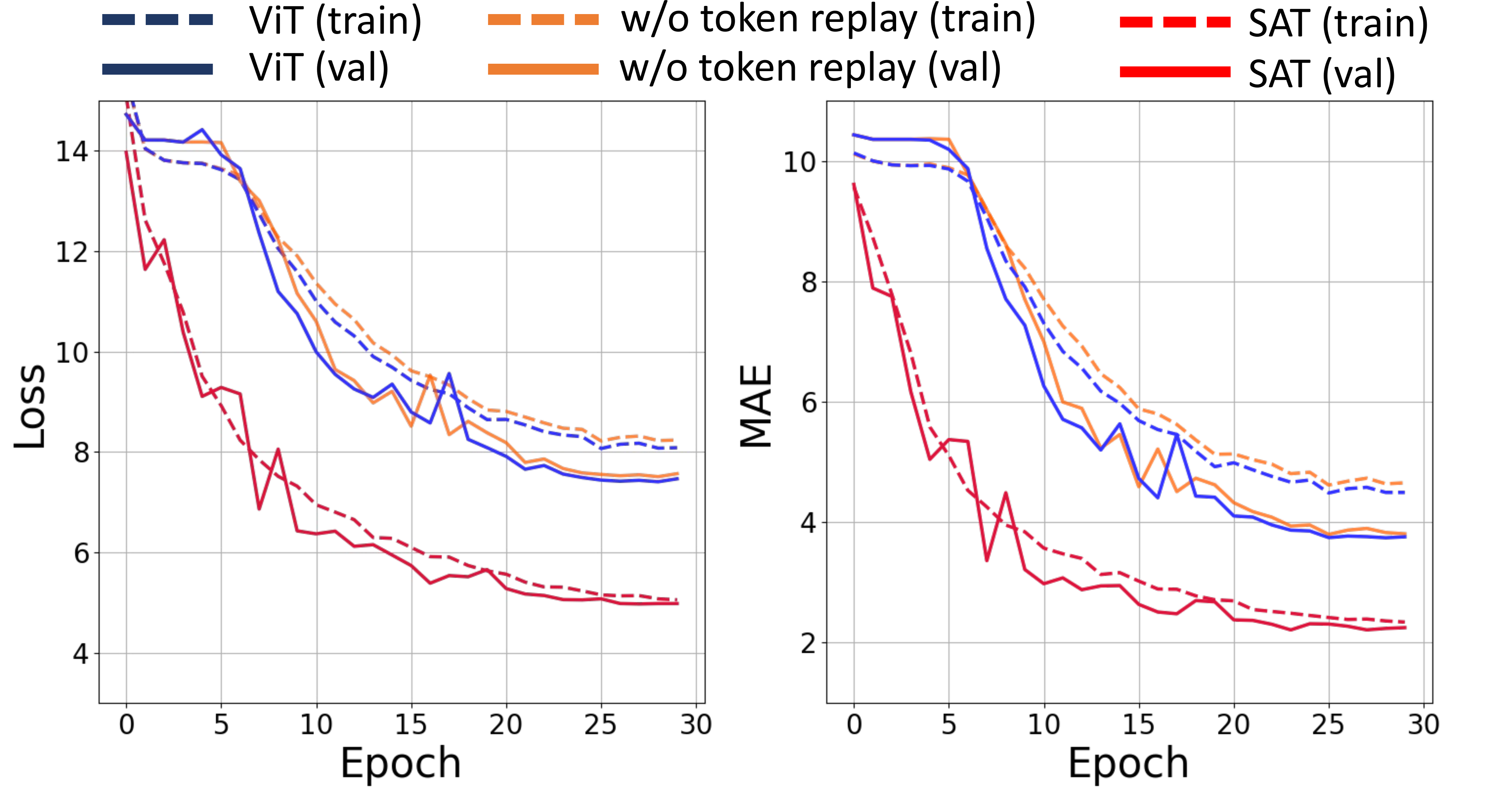}\hspace{0mm}} 
    \caption{(a) Relevance maps \cite{transformer_interpretability} visualizing attention between [CLS] and regional token (Left: vanilla ViT, Right: SAT). 
    (b) Effect of token replay on optimization and generalization (Left: loss, Right: mean absolute error).
    }
\end{figure*}

When training the vanilla ViT as described above, we have observed that the attention module emphasizes excessively on inter-RoI patches.
As shown in Fig. ~\ref{fig:relevance}, olecranon has gained most of the attention from inter RoIs.
Indeed, this result can be interpreted as natural behavior when assessing bone age, as an interpretation of bone age using olecranon alone is the simple yet effective method in clinical practice \cite{dimeglio2005accuracy}.
However, to obtain a better elaborate and accurate interpretation of bone age, scores from all RoIs based on their morphology is essential to be obtained in the Sauvegrain method ~\cite{dimeglio2005accuracy}.
Thus, true correlations between maturity scores are better to be more isotropic, i.e. discrepancies are allowed for inter-landmarks but predictions on a single landmark must be identical.
To resolve the disparity between ViT and expected behaviors, we propose two modifications: (1) \textit{token replay} method that repeatedly adds patch embeddings to intermediary features with corresponding tokens, and (2) RAB, which explicitly imposes regional bias on the attention map.

\subsection{Self-accumulative Vision Transformer}
\smalltitle{Token Replay}
Regional predictions (classification heads) do not necessarily prioritize their corresponding regional tokens.
Since each class token in the last encoder layer that is used for computing the each maturity score utilizes the self-attention mechanism with multiple regional tokens and other class tokens, their own region-specific semantic information could be hindered and mixed with other tokens. Thus, they could not preserve their own region-specific information.
In contrast, we argue that intentionally considered isotropic behavior could be efficient to improve the classification performance.
Therefore \textit{token replay} is designed to preserve region-specific signals in predicting maturity scores by ``replaying'' input regional tokens $z_0 = \left[e_1, \dots, e_R\right]$ so that $z_0$ are added to \verb|[CLS]| tokens encoding as $\tilde{z}_l = h_l \left(\left[\tilde{z}_{l-1}, z_{l-1}\right]\right) + \left[z_0, 0\right]$ at each layer $l$.
Fig. \ref{fig:dongbin_figure_5} illustrates how token replay works in each encoder layer.

Take the final features $\tilde{z}_L + z_0$ (with a slight abuse of notation) as an example used by classifier heads to predict maturity stages.
Similar to how information propagation is improved through the use of residual connections in neural networks \cite{resnet,densenet}, classification heads are guaranteed to attribute weights to regional patches.
Fig. \ref{fig:learning_curve} illustrates how token replay improves optimization and generalization, confirming the need that each \verb|[CLS]| tokens should be accentuated by intra-regional signals.

\smalltitle{Regional Attention Bias (RAB)}
Attention modules are designed to underscore more relevant query-key token pairs.
Recall the expected behavior of the maturity score prediction model is to attribute intra-regional features.
Regional predictions in attention modules as-is are not incited to prioritize their corresponding regional tokens.
Consequently, the class-regional token attention relevance scores \cite{transformer_interpretability} are observed in left Fig. \ref{fig:relevance} to be highly anisotropic conflicting with their desired behavior.

To remedy this anisotropic behavior, as shown in Fig. \ref{fig:dongbin_figure_6}, we explicitly add a matrix $B$ to each attention $\tilde{A}$ where $\tilde{A} \in \mathbb{R}^{2R \times 2R}$ denote the pre-softmax attention at a fixed layer and $B$ is a $2 \times 2$ \textit{block matrix} with $0$s on all blocks other than the second block $D_R$.
Thus, the top-right side of the matrix $B$ has values of $D_R$ which denotes the $R$-dimensional diagonal matrix.
RAB $d_r$ in $D_R$ is computed by the following equation
\begin{equation}\label{eq:rab}
    d_r = \tanh{(b_r + 1)} * 0.5.
\end{equation}
Here $b_r$ denotes the learnable scalar for each region.
By amplifying both forward and back-propagation with RAB from the beginning, intra-regional attention is emphasized throughout training. Thus, the attention to intra-landmarks is increasingly emphasized with SAT as presented in Fig. \ref{fig:relevance}.

\section{Experiments}\label{sec:expeirment}
\subsection{Environments}
\smalltitle{Dataset}
Elbow AP and lateral pairs of radiographs were collected from three anonymous tertiary hospitals \footnote{Data sources are currently undergoing disclosure procedures, then will be revealed.}.
The dataset consists of 4,615 radiograph pairs of AP and lateral view positions collected from the two institutions are used for training and validation.
In addition, 164 pairs of AP and lateral radiograph images collected from another different institution were used for the test dataset.
Three and five researchers with more than 7 years of experience labeled the data providing the maturity score of each landmark for each dataset respectively (e.g., lateral condyle, trochlea, olecranon, proximal).
Lastly, landmarks have been extracted from each AP and lateral view elbow radiographs with a key point detection network as used in previous work.\cite{ahn2021assessment}.
Each extracted landmark is resized by $384 \times 384$.
More detailed statistics for the dataset are represented in the table in the supplementary material.

\smalltitle{Baseline algorithms}
Due to lack of Sauvegrain related works, we compare the previous work ~\cite{ahn2021assessment} and its variants to evaluate the BAA performance of SAT: an ensemble of $R$ single-view single-task CNNs (SV-ST CNNs), multi-view single-task CNNs (MV-ST CNNs), and multi-view multi-task CNN (MV-MT CNN), an ensemble of $R$ multi-view single-task ViTs (MV-ST ViTs) and multi-view multi-task ViT (MV-MT ViT).

\smalltitle{Training details}
The CNN used in ViT to embed the token is ResNet18~\cite{resnet}.
We set the depth to 12 with 6 heads and the embedding dimension to 384 for training the SAT and its variants. 
Models were trained using the SAM optimizer \cite{sam} with cosine annealing on a batch size of $16$ for 30 epochs.
The initial learning rate is 0.01.
For the data augmentation, we have rotated the image randomly between ±15°, shifted randomly to ±32 pixels, and flipped horizontally \footnote{Github repository URL will be updated after the review.}.

\begin{table*}[t]
\centering
\caption{Comparison results on the MAE for each landmark score, summation of the score, and skeletal age on
\textbf{test dataset}. TR refers to token replay. 
Wilcoxon $t$-test is performed between SAT and others on BAA. ($^{*}$: $P$-value<0.05; $^{**}$: $P$-value<0.01; $^{***}$: $P$-value<0.001.).
}
\label{tab:baa-result}
\begin{tabular}{c|c||c|c|c|c|c|c|c}
\toprule
Method &$\#$  & Lat. & Troch. & Prox. & Olec. & Prox. & Sum 
& BAA \\
& params & cond. &  & (AP) &  & (Lat) & 
& 
\\
\midrule
SAT (ours) & 33M &
\textbf{0.354} &
0.190 & 0.276 & \textbf{0.169} & 0.308
& \textbf{0.709} & \textbf{0.261} \\
\hline
\hline
SV-ST CNNs~\cite{ahn2021assessment} & 90M
& 0.455 & 0.301 & 0.398
& 0.252 & 0.372 & 1.054 & 0.372$^{***}$ \\
MV-ST CNNs & 90M
& 0.411 & 0.331 & 0.336
& 0.239 & 0.330 & 0.915 & 0.334$^{***}$ \\
MV-MT CNN & 18M
& 0.422 & 0.282 & 0.457
& 0.265 & 0.451 & 1.032 & 0.381$^{***}$ \\
MV-ST ViTs & 168M
& 0.357 & 0.210 & \textbf{0.256}
& 0.192 &\textbf{0.257} & 0.728 & 0.263
\\
MV-MT ViT & 33M
& 0.451 & 0.376 & 0.288
& 0.330 & 0.288 & 0.922 & 0.324$^{***}$
\\
\hline
\hline
SAT w/o TR & 33M & 0.439 & 0.341 & 0.279
& 0.321 & 0.289 & 0.888 & 0.318$^{***}$
\\
SAT w/o RAB & 33M & 0.356 & \textbf{0.189} & 0.281
& 0.176 & 0.299 & 0.720 & 0.261
\\
\bottomrule
\end{tabular}
\end{table*}

\subsection{Experimental Results}
\smalltitle{Performance Comparison}
Table \ref{tab:baa-result} presents the number of parameters and mean absolute error (MAE) of each landmark, their sum, and their conversion to BAA.
The performance of MAE has been evaluated on the test set using five-fold cross-validation on the training set (80$\%$ training and 20$\%$ validation). 
The SAT model outperforms other compared methods, achieving 0.261 MAE on BAA.
Our results showed that the multi-view strategy is effective in both CNN-based and ViT-based architectures for solving the Sauvegrain method.
However, the multi-task problem remains challenging, as demonstrated by the inferior performance of MV-MT CNN and MV-MT ViT compared to MV-ST CNNs and MV-ST ViTs, respectively. 
It could be interpreted as multi-task methods were detrimentally affected by over-reliance on inter-landmark inputs.
Interestingly, our SAT model even outperformed MV-ST ViTs with significantly fewer parameters, suggesting that SAT explicitly emphasizes the attention of landmarks isotropically and benefits from both multi-view information and the interplay between landmarks, while addressing the multi-task problem.
Comparison using cumulative score~\cite{cumulative_score2,cumulative_score} is reported in the supplementary section.

\smalltitle{Ablation Study}
The last two rows in Table \ref{tab:baa-result} show the influence of each SAT component.
We have removed each component of our SAT and reported the BAA results.
Compared with MV-MT ViT, both \textit{token replay} and \textit{RAB} are crucial for the performance, however, the former contribute the most when combined together as shown in Table \ref{tab:baa-result}.
Nonetheless, utilizing RAB consistently outperforms ViT and induces higher relevance maps in its attention which is desirable in understanding what the classifier attributes its predictions to.

\begin{table*}
  \centering
     \caption{Comparison results on BAA differing encoder of SAT. Wilcoxon $t$-test is performed between ResNet and others, but there is no statistical significance.}
    \label{tab:encode_ablation}
    \begin{tabular}{c||c|c|c||c|c}
    \toprule
        Encoder & $\#$ & Inference time &  Training  & Sum & BAA \\
       & Params & (CPU/GPU) & time & \\
        \midrule
            ResNet18\cite{resnet} & 33M & 
            0.21s / 0.03s & 2.7h & 0.709 & 0.261 \\
            VGG16 \cite{vgg} &  157M &
                1.13s / 0.06s  & 10h & 0.674 & 0.248 \\
            Densenet121 \cite{densenet} & 29M &
                0.98s / 0.05s  & 6h & 0.708 &  0.260 \\
            HR-Net-w18-small\cite{hrnet} & 33M & 
                0.26s / 0.04s  & 3.7h & 0.710 & 0.261 \\
            ResNext50-32x4d \cite{resnext} & 44M &
                0.60s / 0.06s  & 8.5h & 0.699 & 0.252 \\
        \bottomrule
      \end{tabular}
\end{table*} 

 \begin{figure*}[t]
	\centering 
{\includegraphics[width=0.9\columnwidth]{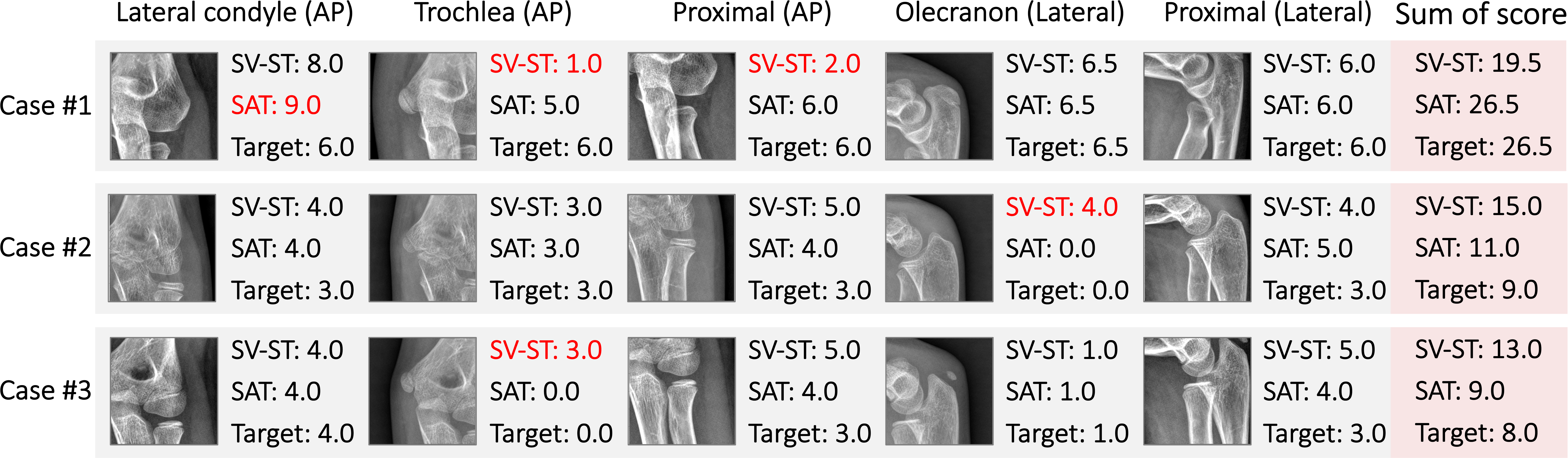}\hspace{0mm}}
	\caption{Three cases of prediction results on the single-view single-task (SV-ST) CNNs and SAT with ground truth score. The red color highlights the prediction result where the difference from the ground truth is more than 3.}
	\label{fig:showcases}
\end{figure*}

Comparisons between other CNN encoders are reported in Table \ref{tab:encode_ablation}. 
We have observed that ResNet-18 shows the most efficient computational cost.
Though other models, such as VGG and ResNext, shows better BAA results than ResNet-18, their computational cost is not efficient as ResNet.
There is no statistically significant difference between the ResNet-18 and other models.
Thus, we have used the baseline encoder of SAT for ResNet-18 which has the least number of parameters.

\smalltitle{Case Analysis}
To verify that SAT indeed has lower variance in predicting the maturity score in real data, we have analyzed the 3 cases where SV-ST CNNs \cite{ahn2021assessment} has shown the largest MAE in summation of scores in Fig. \ref{fig:showcases}. Prediction score for SV-ST CNNs and SAT is reported with ground truth score.
As shown in Fig \ref{fig:showcases}, not only the prediction error of SV-ST CNNs but also the variance in the scores is much bigger than SAT does.
It demonstrates that SV-ST CNNs have limitations when applied to Sauvegrain methods, where the maturity score of each RoI is highly correlated.
On the other hand, SAT shows better prediction on each RoI and lower variance than SV-ST CNNs, demonstrating that SAT could be a practical solution for the hard cases.

\section{Conclusion}\label{sec:conclusion}
This work studied the Sauvegrain-based BAA and identified issues with DNNs trained for the MV-MT ordinal classification problem. Ensembling CNNs increases computational costs and vanilla ViT leads to anisotropic attention and prediction discrepancies. To address these issues, this work introduced SAT, consisting of token replay and regional attention bias techniques, which were effective in mitigating these problems. This approach has broader implications for training ViT for MV-MT ordinal classification. Applied to Sauvegrain-based BAA, SAT is clinically meaningful in assisting diagnosis of precocious and delayed maturity in adolescents.
\newpage

\bibliographystyle{splncs04}
\bibliography{egbib}
\newpage

\renewcommand{\thetable}{S\arabic{table}}  
\renewcommand{\thefigure}{S\arabic{figure}}

% \title{Supplementary Material}

% \maketitle
% \begin{document}
\section*{Supplementary Material}\label{sec:supple}

\begin{figure*}[h]
    \centering
    \subfigure[\label{fig:dongbin_1_fig:b}SV-ST]{\includegraphics[width=0.2\columnwidth]{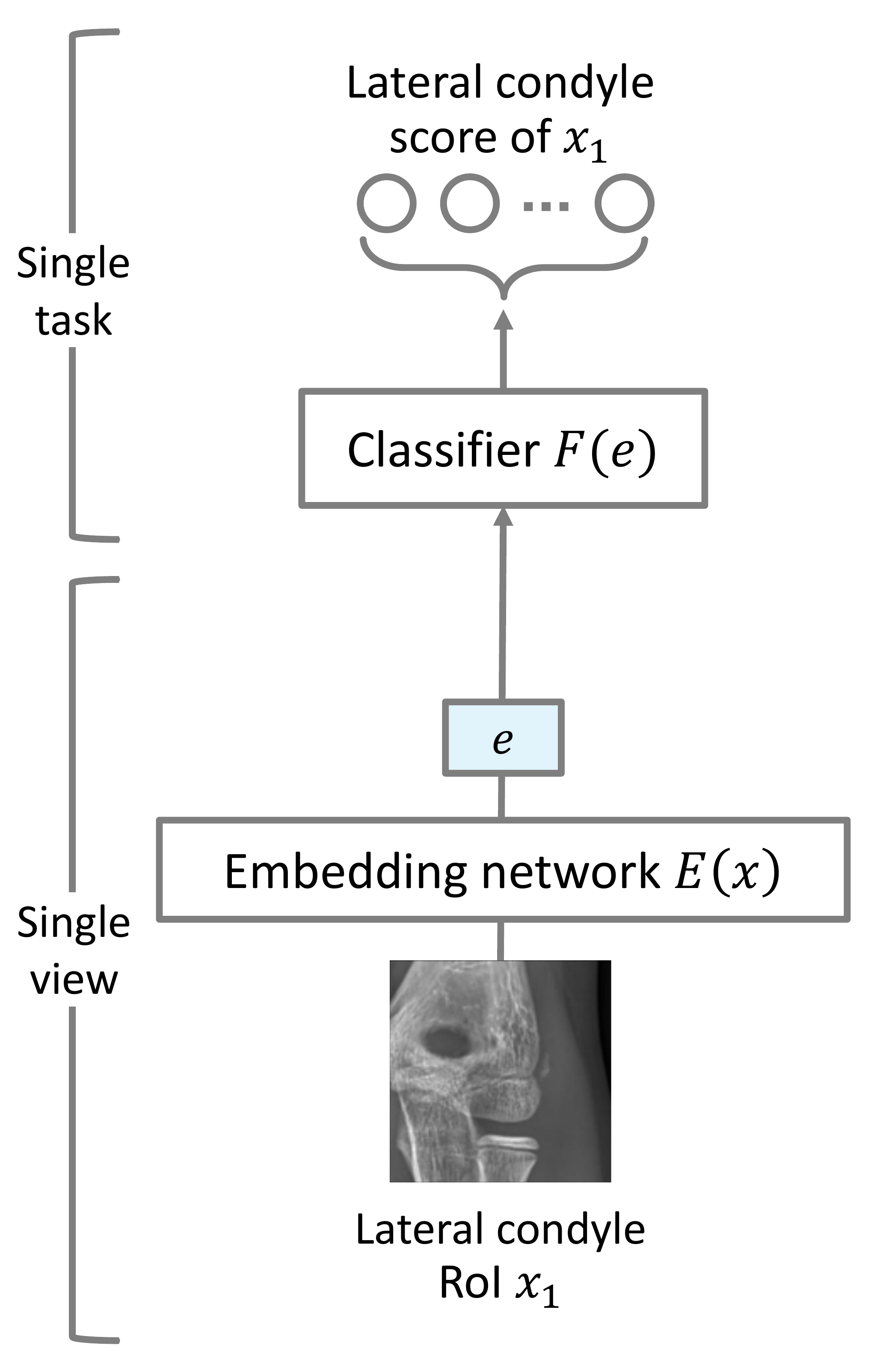}\hspace{1mm}} 
    \subfigure[\label{fig:dongbin_1_fig:a}MV-ST]{\hspace{1mm}\includegraphics[width=0.36\columnwidth]{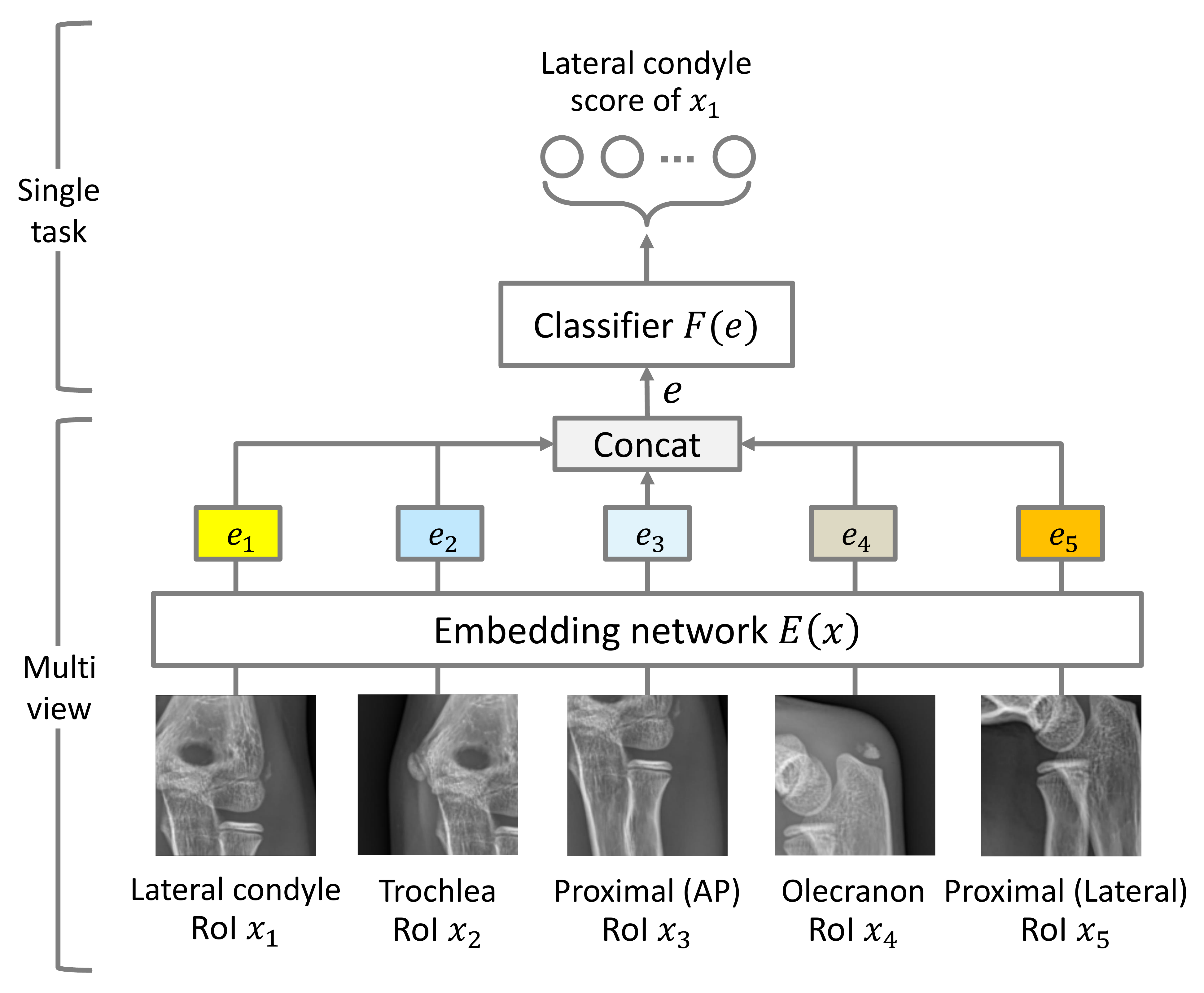}\hspace{1mm}}
    \subfigure[\label{fig:dongbin_1_fig:c}MV-MT]{\hspace{1mm}\includegraphics[width=0.36\columnwidth]{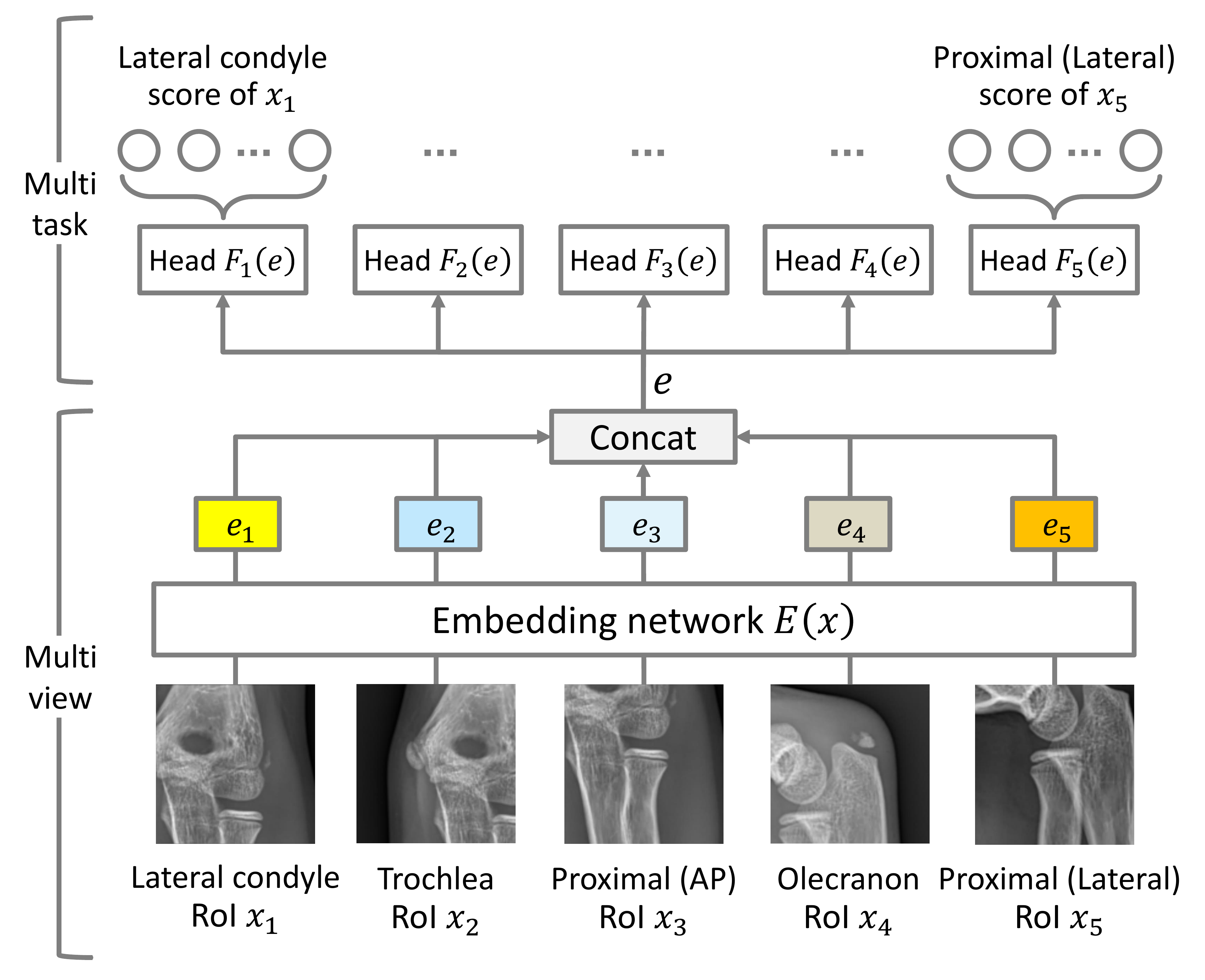}}
    \caption{Comparison of the single-view single task (SV-ST), multi-view single-task (MV-ST), and multi-view multi-task (MV-MT).
    For applying the Sauvegrain method with DNNs, one can utilize the three commonly-used architectures based on the number of views and the number of tasks (SV-ST, MV-ST, MV-MT).}
    \label{fig_supp:}
\end{figure*}

\begin{table}[]
  \centering
     \caption{Statistics of training, validation, and test dataset. M/F in \textit{Sex} ratio refers to the male and female. From \textit{Lateral condyle} to \textit{Proximal}, the numbers in the table refer to the maturity score of each RoI. Numbers in \textit{Bone age} refer to the years. In addition, numbers in between \textit{Lateral condyle} and \textit{Bone age} is following the format of mean $\pm$ standard deviation.}\label{tab:data-statistic}
    \begin{tabular}{c|c|c}
    \toprule
        & Train / Validation & Test \\
        \midrule
        Number of data & 4,615 & 164 \\
        Sex ratio (M/F) & 0.71 / 0.29  & 0.51 / 0.49 \\
        Lateral condyle & 6.32$\pm$2.85 & 6.31$\pm$2.44   \\
        Trochlea &  2.97$\pm$2.16 & 2.9$\pm$1.99 \\ 
        Olecranon  &   4.65$\pm$1.79 &  4.71$\pm$1.26 \\
        Proximal & 3.98$\pm$3.03 &  3.98$\pm$2.79 \\ 
        Sum of score & 17.92$\pm$9.38  & 17.91$\pm$8.18 \\ 
        Bone age & 11.55$\pm$3.88 & 11.56$\pm$1.87   \\
        \bottomrule
      \end{tabular}
\end{table}

\begin{table}[h]
\caption{Training details for SAT. For fair comparison other ViT-based models (e.g. MV-MT ViT, MV-ST ViTs) 
 follow this setting.}\label{tab:data-statistic}
  \centering{
    \begin{tabular}{c|c|c|c|c|c}
    \toprule
         & Epochs & Batch size & Learning rate & Optimizer & Learning rate scheduler \\
        \midrule
        \hline
        SAT & 16 &  30 & 0.01 & SAM \cite{sam} & cosine annealing \\
        \bottomrule
      \end{tabular}
  }
  
\end{table}
\begin{table}[]
  \centering
     \caption{Evaluation metric for comparing the performance of SAT and others. $y_i$,$\hat{y}_i$ deontes the ground truth and predicted value respectively. $\theta$ is the hyperparameter where it decides that the difference value between prediction and ground truth label belongs to a certain range.}\label{tab:data-statistic}
    \begin{tabular}{c|c}
    \toprule
        metric & equation \\
        \midrule
        \hline
        MAE (Mean Absolute Error)
       & $\frac{1}{N} \sum_{i=1}^{N}|y_i - \hat{y}_i|$ \\
       \hline
        CS{($\theta$)}
       & $\frac{1}{N} \sum_{i=1}^{N} \mathbbm{1}(|y_i - \hat{y}_i| \leq \theta) \times 100\%$ \\
        \bottomrule
      \end{tabular}
\end{table}

\begin{figure*}[h]
	\centering 
    \includegraphics[width=0.48\columnwidth]{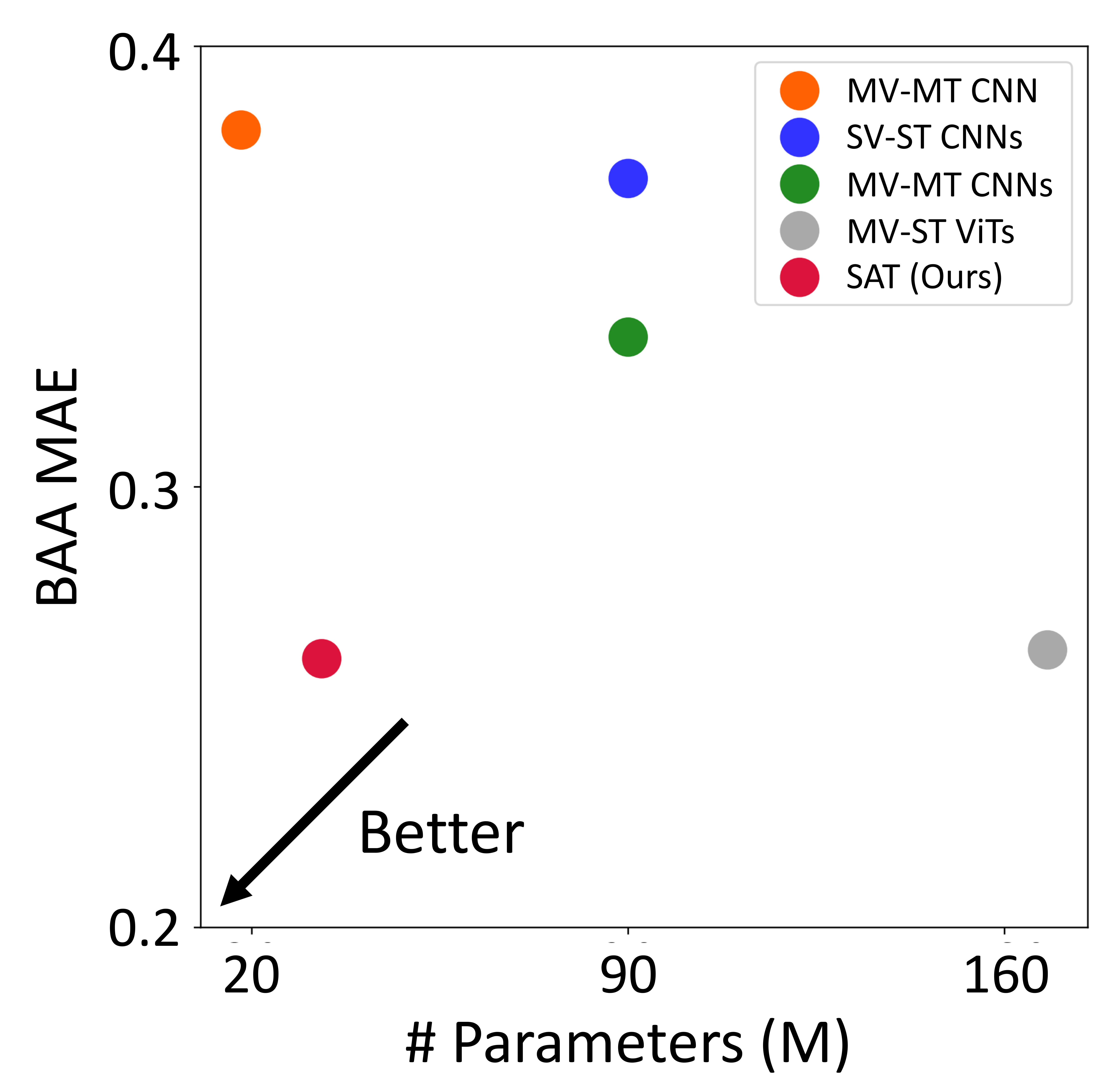}
	\caption{Comparison of performance based on the BAA and the number of parameters on the methods used in experiments.}
	\label{fig:fig_mae_params}
\end{figure*}

\begin{table*}[t]
  \centering
     \caption{Comparison results on cumulative score (\textbf{$CS_{(\theta=1)}$}) for each landmark. We set the age range $\theta$ to 1. The higher is better. Results are represented as mean ± standard deviation (SD).}
    \label{tab:baa-result-accuracy}
    \begin{tabular}{c||c|c|c|c|c||c}
    \toprule
        Method &  Lat. cond. & Troch. & Prox. (AP) & Olec. & Prox. (Lat) & Sum \\
        \midrule
            SAT (ours) & 
                 97.56$\pm{0.67}$ & 98.78$\pm{0.39}$ & 98.05$\pm{0.46}$ & 96.46$\pm{0.46}$  & 96.34$\pm{0.86}$
                  & 81.83$\pm{1.74}$ \\
                  \hline
                  \hline
            SV-ST CNNs & 
                94.63$\pm{1.12}$  & 95.61$\pm{1.61}$ & 94.63$\pm{2.02}$
                & 95.37$\pm{0.99}$  & 95.98$\pm{1.47}$ & 72.44$\pm{4.05}$ \\
            MV-ST CNNs & 
                 96.83$\pm{1.51}$ & 95.24$\pm{1.51}$ & 95.73$\pm{0.86}$ &95.37$\pm{2.46}$ 
                  &  95.73$\pm{1.39}$ & 75.12$\pm{5.58}$
            \\
            MV-MT CNN & 
                 94.88$\pm{1.47}$ & 95.61$\pm{1.61}$  & 90.49$\pm{0.99}$ & 93.42$\pm{1.91}$ 
                  & 91.59$\pm{2.42}$ & 73.17$\pm{2.04}$
            \\
            MV-ST ViTs & 97.56$\pm{0.55}$ &
                  98.90$\pm{0.71}$ & 98.78$\pm{0.00}$ & 96.95$\pm{0.39}$ & 98.78$\pm{0.39}$ & 81.22$\pm{2.48}$
                \\
            MV-MT ViT & 93.42$\pm{1.18}$ &
                  93.29$\pm{0.55}$ & 96.71$\pm{0.30}$ & 94.39$\pm{0.46}$ & 96.71$\pm{0.30}$ & 74.51$\pm{1.51}$
                  \\
        \bottomrule
      \end{tabular}
\end{table*}

% \end{document}

\end{document}